\documentclass[conference]{IEEEtran}
\IEEEoverridecommandlockouts
\usepackage{cite}
\usepackage{amsmath,amssymb,amsfonts}
\usepackage{algorithmic}
\usepackage{graphicx}
\usepackage{textcomp}
\usepackage{xcolor}
\usepackage{float}
\usepackage{booktabs}
\usepackage{tabularx}
\usepackage{placeins}

\def\BibTeX{{\rm B\kern-.05em{\sc i\kern-.025em b}\kern-.08em
    T\kern-.1667em\lower.7ex\hbox{E}\kern-.125emX}}
\begin{document}

\IEEEoverridecommandlockouts \IEEEpubid{\makebox[\columnwidth]{979-8-3503-7943-3/24/\$31.00~\copyright2024 IEEE\hfill} \hspace{\columnsep}\makebox[\columnwidth]{ }}

\title{Scaling BERT Models for Turkish Automatic Punctuation and Capitalization Correction}

\author{
\IEEEauthorblockN{Abdulkader Saoud}
\IEEEauthorblockA{\textit{Department of Computer Engineering} \\
\textit{Yildiz Technical University}\\
Istanbul, Turkey \\
abdulkader.saoud@std.yildiz.edu.tr}
\and
\IEEEauthorblockN{Mahmut Alomeyr}
\IEEEauthorblockA{\textit{Department of Computer Engineering} \\
\textit{Yildiz Technical University}\\
Istanbul, Turkey \\
mahumut.alomeyr@std.yildiz.edu.tr}
\IEEEauthorblockN{Himmet Toprak Kesgin}
\IEEEauthorblockA{\textit{Department of Computer Engineering} \\
\textit{Yildiz Technical University}\\
Istanbul, Turkey \\
tkesgin@yildiz.edu.tr}
\and
\IEEEauthorblockN{Mehmet Fatih Amasyali}
\IEEEauthorblockA{\textit{Department of Computer Engineering} \\
\textit{Yildiz Technical University}\\
Istanbul, Turkey \\
amasyali@yildiz.edu.tr}}



\maketitle


\begin{abstract}
This paper investigates the effectiveness of BERT-based models for automated punctuation and capitalization corrections in Turkish texts across five distinct model sizes. The models are designated as Tiny, Mini, Small, Medium, and Base. The design and capabilities of each model are tailored to address the specific challenges of the Turkish language, with a focus on optimizing performance while minimizing computational overhead. The study presents a systematic comparison of the performance metrics—precision, recall, and F1 score—of each model, offering insights into their applicability in diverse operational contexts. The results demonstrate a significant improvement in text readability and accuracy as model size increases, with the Base model achieving the highest correction precision. This research provides a comprehensive guide for selecting the appropriate model size based on specific user needs and computational resources, establishing a framework for deploying these models in real-world applications to enhance the quality of written Turkish.
\end{abstract}


\section{Introduction}
Clear communication is critical to successful interactions, and poor communication can lead to misunderstandings.
There are two main ways to communicate: verbally and in writing.
Written communication requires punctuation.
These punctuation marks significantly affect how sentences are interpreted.
This is especially important in Turkish, which is characterized by its detailed structure, where proper punctuation is necessary to avoid confusion and clearly convey the intended message.

Despite the critical role of punctuation and capitalization in maintaining textual integrity, current natural language processing (NLP) tools often fall short, especially for languages like Turkish that have unique linguistic features. Existing models are predominantly trained on English language datasets, which do not sufficiently capture the syntactic and grammatical nuances of Turkish. This discrepancy often results in suboptimal performance when these models are applied to Turkish texts, potentially distorting the intended meaning of the communication. Moreover, there is a notable lack of resources that specifically address the automatic correction of punctuation and capitalization in Turkish, creating a significant gap in NLP applications for Turkish speakers.

This paper presents a model that not only corrects punctuation and capitalization errors more effectively, but also makes clear, error-free written communication available to more people. By improving the accuracy of punctuation and capitalization, the model also improves the performance of text-to-speech (TTS) systems, making the spoken version of written texts sound more natural and easier to understand.In addition, this improved model can be a crucial tool for proofreading and editing, significantly improving the quality of Turkish writing on various platforms, from academic articles to social media content. Ultimately, this innovation aims to improve overall written communication in Turkish, ensuring clear and accurate information exchange in both informal and formal settings.

In this research, we used the "batubayk/TR-News"\cite{dataset_article} dataset, accessible through the Hugging Face\cite{huggingface} platform, which contains a wide range of Turkish news articles. The dataset was carefully segmented into sub-paragraphs, each limited to 512 tokens, in order to optimize the training process of the BERT models. Each token was then classified into different classes, paving the way for the fine-tuning phase. We used several variants of the "ytu-ce-cosmos" pre-trained BERT models\cite{toprak2023developing}-from tiny to base-to explore the effect of model size on performance. Parameters were carefully adjusted to optimize accuracy, and performance comparisons were made between the different model sizes to determine the most effective configuration for punctuation and capitalization tasks in Turkish texts.

Following the fine-tuning phase, we conducted a series of experiments to evaluate the performance of our models. This included rigorous testing under various conditions to ensure the robustness and reliability of the punctuation and capitalization predictions. The evaluation metrics focused on precision, recall, and F1 score, which are critical in assessing the effectiveness of NLP models.
These experiments not only demonstrated the superior performance of our models over existing solutions, but also highlighted their practical applications in real-world scenarios, such as improving the readability of text and supporting automated content correction.

\section{Literature Review}

The proposed method is based on the classification of individual tokens within a sentence, rather than the entire sentence or document, as is the case with sentiment analysis or topic classification. This system, referred to as Token Classification, is frequently employed for Named Entity Recognition (NER) task\cite{deneme}. However, it can be utilized in a self-supervised manner by labeling each token with the punctuation mark that follows it or whether the word begins with a capital letter.

Various methods have been explored to correct punctuation errors and predict the use of capital letters. Token Classification is the most commonly used method in this domain. CRFs (Conditional Random Fields) are generally used algorithms for Token Classification \cite{CRF}.

Traditional methods often combine prosodic and lexical features using recurrent neural networks, while more recent approaches leverage Transformer-based models. The state-of-the-art method proposed by Courtland et al. \cite{courtland2020efficient} employs a BERT-based model that achieves high accuracy. This study contributes to the existing literature by developing BERT-based punctuation restoration models for English and Hungarian. The results demonstrate significant advancements in this field.

In recent research, Ling et al. have introduced a fast and compact BERT model specifically designed for punctuation correction in Chinese texts \cite{ling2023small}. This model retains 95\% of the performance capabilities of models ten times its size, although its application has been limited solely to the medical field."

In their research, Hieu and Dinh \cite{Viet} demonstrated the superiority of monolingual models over multilingual counterparts in accurately predicting punctuation in Vietnamese texts. Their findings underscore the efficacy of tailored, language-specific models in handling the nuanced aspects of punctuation. Furthermore, they highlighted the potential of transfer learning in enhancing model performance. Building on these insights, they proposed an augmented approach that incorporates both Bidirectional Long Short-Term Memory (Bi-LSTM) layers and CRF to further refine the accuracy of punctuation prediction. This method leverages the strengths of deep learning and sequence modeling to optimize text processing tasks.

One effective method in NLP is using transformers. A study \cite{soudArticle} tested this approach on English and Bengali, achieving excellent results for English and providing initial insights for Bengali. Another study \cite{mahmoudArtile} focused on Turkish, creating a specific dataset for this language. They applied models such as BERT \cite{bert}, ELECTRA \cite{electra}, and ConvBERT \cite{convbert} to see how well they could handle Turkish text. It should be noted, however, that this study is trained on a very small number of punctuation marks (only three in total) and is limited to the prediction of punctuation marks. These studies show the adaptability of transformer models to different languages.

In contrast to the models presented in the literature, our model is specifically designed to predict and correct eight different punctuation marks, rather than the three punctuation marks typically addressed in previous models. Additionally, it is capable of not only predicting punctuation but also determining whether a word is written entirely in lowercase, uppercase, or first letter uppercase. Furthermore, it offers five distinct models for speed and performance, including models that are both highly efficient and compact, as well as larger and more successful models.

\section{Method}
 This section details the methodologies employed in this project to develop and evaluate Turkish BERT models for NER and punctuation prediction tasks. It covers the models and architectures used, data preprocessing and training procedures, and the evaluation metrics applied to assess the models' performance. The following subsections provide an in-depth overview of these components.   

 \subsection{Models and Architectures}
        In this paper, BERT models trained by YTU-CE-COSMOS have been utilized. These models are Tiny, Mini, Small, Medium, and Base, and the differences, advantages, and disadvantages of each have been discussed in detail.    
        BERT is a deep learning model based on transformer architecture and works with bidirectional logic on texts. The BERT model is used for various NLP tasks such as Question Answering, Mask Filling, Text Classification, and more.
        The BERT model consists of two main parts:
        \begin{itemize}
            \item \textbf{Encoder:} The set of layers where the text is taken as input and the tokens are represented.
            \item \textbf{Output Layer:} The layer where outputs for customized tasks are produced. This layer has been adjusted specifically for the NER task of the project.
        \end{itemize}
        The characteristics of the Turkish BERT models (tiny, mini, small, medium, and base) have been examined in detail. Architectural details such as hidden dimensions, number of attention heads, hidden layers, and total parameters are presented in the Table \textbf{\ref{mylabel}}.
        
        \begin{table}[h]
            \caption{Architectural differences and parameter counts among the models.}
            \centering
            \small
            \begin{tabularx}{\columnwidth}{@{}l*{4}{X}c@{}}
                \toprule
                \textbf{Model} & \textbf{Hidden Size} & \textbf{Num. Attn Heads} & \textbf{Num. Hidden Layers} & \textbf{Num. Parameters (millions)} \\
                \midrule
                \textbf{Tiny}   & 128  & 2  & 2  & 4.6 \\
                \textbf{Mini}   & 256  & 4  & 4  & 11.6 \\
                \textbf{Small}  & 512  & 8  & 4  & 29.6 \\
                \textbf{Medium} & 512  & 8  & 8  & 42.2 \\
                \textbf{Base}   & 768  & 12 & 12 & 110.7 \\
                \bottomrule
            \end{tabularx}
            \label{mylabel}
        \end{table}

        As we can see from the Table \textbf{\ref{mylabel}}, the model size increases significantly with larger models. While larger models may yield better results, they consume a substantial amount of resources compared to smaller ones.

        The success of the model during the training phase depends on the correct setting of parameters. For training, AdamW \cite{adamW} with scheduled weight decay was tested with different learning rates and batch sizes.

        \subsection{Data Preprocessing and Training}
        For this project, the 'batubayk/TR-News'\cite{dataset_article} dataset , which is 760 MB in size, was used. The selected dataset contains clean news sentences and the news texts usually use correct punctuation, making it ideal for our study. The dataset consists of 9 columns, but only the 'content' column, which consists solely of paragraphs, was used, and the other columns were deleted. The distribution of the dataset into train, test, and validation sections was done as 70\%, 20\%, and 10\% respectively. The distribution of punctuation marks can be seen in Table \ref{tab:noktalama}.
        \begin{table}[h]
            \caption{Distribution of Punctuation Marks}
            \centering
                \begin{tabular}{|c|c|c|c|}
                \hline
                Split & Train      & Test    & Validation \\ \hline
                .     & 3.686.111  & 1.069.011 & 524.369    \\ \hline
                ,     & 3.317.388  & 948.878 & 473.973    \\ \hline
                !     & 15.096    &  4.168  & 2.002     \\ \hline
                ?     & 71.326     & 20.439   & 9.832      \\ \hline
                ;     & 56.987     & 16.707   & 7.904      \\ \hline
                :     & 151.911    & 42.898  & 22.678     \\ \hline
                -     & 276.109    & 78.669  & 38.660     \\ \hline
                '     & 2.257.317  & 647.035 & 323.570    \\ \hline
            \end{tabular}
            
            \label{tab:noktalama}
        \end{table}

        Since the tokenizer is uncased (accepting only lowercase letters), uppercase letters were converted to lowercase. Additionally, because the tokenizer is designed with a 512 input sequence limit, long paragraphs were divided into parts. The first segment includes 512 tokens starting from the beginning; subsequent segments begin immediately following the last punctuation mark (period, exclamation mark, semicolon, or question mark) of the previous segment.

        For the punctuation prediction model, after extracting the tokens, the label assignment phase is carried out. In NER training, it is necessary to assign a label to each token. In the project, there are 9 labels: '.', ',', '!', '?', ';', ':', '-', "'" and 'non'. If a word is followed by a punctuation mark, the appropriate label for that mark is assigned; otherwise, the 'non' label is assigned. Finally, punctuation marks are removed from the token sequence.

        \begin{table}[h]
            \caption{Sample Punctuation Preprocessing}
            \centering
            \begin{tabular}{|p{3.5cm}|p{4.5cm}|}
            \hline
            \textbf{Sentence}  & Türkiye'nin her tarafında devam etmektedir.                    \\ \hline
            \textbf{Token}  & 'türkiye', 'nin', 'her', 'tarafında', 'devam', 'etmektedir' \\ \hline
            \textbf{Label} & 'apostrophe', 'non', 'non', 'non', 'non', 'period'                        \\ \hline
            \end{tabular}
            \label{tab:exampleData}
        \end{table}

        In Table \textbf{\ref{tab:exampleData}}, sample data consisting of tokens and labels for the punctuation prediction model are shown from the original sentence.

        For the capital letter prediction model, a word can begin with a capital letter ("One"), have all its letters capitalized ("CAP"), or be entirely in lowercase ("non"). Each token has been assigned a corresponding label. Sample data is presented in Table \textbf{\ref{tab:exampleDataCap}}.

        \begin{table}[h]
            \caption{Sample Capitalisation Preprocessing}
            \centering
            \begin{tabular}{|p{3.5cm}|p{4.5cm}|}
            \hline
            \textbf{Sentence}  & YTU Türkiye'nin en iyi okuludur.                    \\ \hline
            \textbf{Token}  & 'y', '\#\#tu', 'türkiye', 'nin', 'en', 'iyi', 'okulu', '\#\#dur' \\ \hline
            \textbf{Label} & 'Cap', 'non', 'One', 'non', 'non', 'non', 'non', 'non'                      \\ \hline
            \end{tabular}
            \label{tab:exampleDataCap}
        \end{table}
        
        To train the model, the AdamW optimizer with a linear scheduler was utilized, testing various learning rates (4e-4 and 5e-5) and batch sizes (128, 32, and 16). In the base model, the gradient clipping technique was applied with the max grad norm=0.5 hyperparameter to ensure model stability. 

        \subsection{Evaluation}
        The metrics used to determine the accuracy of the model's outputs are described below:
        \begin{itemize} 
            \item \textbf{Precision:} Precision denotes the ratio of true positive predictions to the total number of instances predicted as positive by the model. 
            \begin{equation} 
                \text{Precision} = \frac{\text{True Positive}}{\text{True Positive} + \text{False Positive}} 
            \end{equation} 
            \item \textbf{Recall:} This metric indicates the percentage of all actual positive instances that the model correctly predicts as positive.
            \begin{equation}
                \text{Recall} = \frac{\text{True Positive}}{\text{True Positive} + \text{False Negative}}
            \end{equation}
            
            \item \textbf{F1 Score:} The F1 score is the harmonic mean of Precision and Recall, balancing both metrics.
            \begin{equation}
                F1 = 2 \times \frac{\text{Precision} \times \text{Recall}}{\text{Precision} + \text{Recall}}
            \end{equation}
            
            \item \textbf{Confusion Matrix:} The Confusion Matrix is a table that shows the model's true positive, false positive, false negative, and true negative predictions for each class. This matrix is used to visually assess the model's performance in detail.
        \end{itemize}

\section{Performance Analysis}
    
    \subsubsection{Preprocessing}
    The model was initially trained by tokenizing sentences using the NLTK library. This method yielded good results during the testing phase. However, in real-world tests, since the accuracy of the punctuation marks entered by the user is unknown, paragraphs could not be split into sentences. When the model was tested without splitting paragraphs, it did not yield satisfactory results. Therefore, the method described in the system design section, which involves using the paragraph without splitting, was employed.
    
    \subsubsection{the affect of learning rate}
    
    Models were trained and tested using different learning rate values. The results of the tests are shown in Tables \textbf{\ref{tab:lrBH}} and \textbf{\ref{tab:lrN}}. Changing the learning rate value yielded different results for each model. For example, a value of 4e-4 performed better than 5e-5 in all models except for Base. Finally, the modules with the highest F1 scores in the experiments were used.
    
    \begin{table}[h]
        \caption{Capitalization Models learning rates}
        \centering
        \begin{tabular}{|c|c|c|c|c|}
        \hline
        \textbf{Model}  & \textbf{Learning Rate} & \textbf{F1}    & \textbf{Learning Rate} & \textbf{F1}    \\ \hline
        Tiny   & 5e-5          & 0.82  & 4e-4          & 0.86  \\ \hline
        Mini   & 5e-5          & 0.88  & 4e-4          & 0.90  \\ \hline
        Small  & 5e-5          & 0.908 & 4e-4          & 0.918 \\ \hline
        Medium & 5e-5          & 0.916 & 4e-4          & 0.919 \\ \hline
        Base   & 3e-5          & 0.926 & 3e-6          & 0.921 \\ \hline
        \end{tabular}
        
        \label{tab:lrBH}
    \end{table}
    
    \begin{table}[h]
        \caption{Punctuation Models learning rates}
        \centering
        \begin{tabular}{|c|c|c|c|c|}
        
        \hline
        \textbf{Model}  & \textbf{Learning Rate} & \textbf{F1}    & \textbf{Learning Rate} & \textbf{F1}    \\ \hline
        Tiny   & 5e-5          & 0.56 & 4e-4          & 0.65 \\ \hline
        Mini   & 5e-5          & 0.73 & 4e-e          & 0.73 \\ \hline
        Small  & 5e-5          & 0.74 & 4e-4          & 0.76 \\ \hline
        Medium & 5e-5          & 0.76 & 4e-4          & 0.77 \\ \hline
        Base   & 5e-5          & 0.785& 4e-4         & 0.77 \\ \hline
        \end{tabular}
    
        \label{tab:lrN}
    \end{table}

    \subsubsection{Model Testing}
\begin{figure}[htbp]
            \centering
            \includegraphics[scale=0.55]{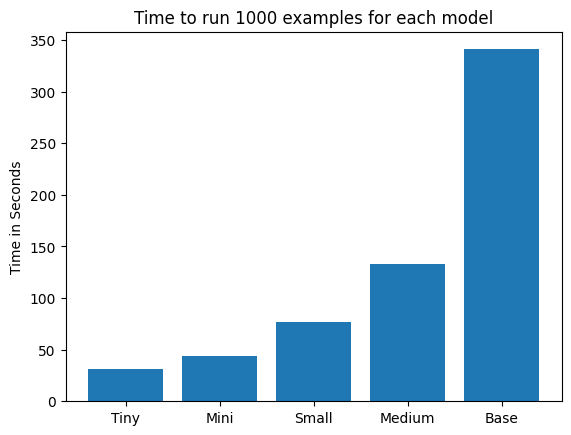}
            \caption{Time to Test 1000 Examples for Each Model}
            \label{fig:TestTime}
        \end{figure}

        To evaluate the models' performance in a real-world scenario, they were tested on a dataset of 1000 examples. This testing phase aimed to assess the models' efficiency and reliability when applied to a substantial amount of data. The time taken to test each model on these examples is summarized in Figure \textbf{\ref{fig:TestTime}}.The tests were conducted using a GPU (NVIDIA GeForce GTX 1060).
        
        As shown in Figure \textbf{\ref{fig:TestTime}}, the time required to test 1000 examples increases significantly with the size of the model. The Tiny model completes the testing in approximately 31 seconds, whereas the Base model takes nearly 6 minutes. This analysis highlights the trade-off between model complexity and testing efficiency, which is crucial for selecting the appropriate model based on resource availability and real-time processing requirements.
        
\section{Experimental Results}
    The F1 scores of the trained models are presented in Table \textbf{\ref{tab:f1Comapre}}.
    
    \begin{table}[H]
        \caption{F1 Scores of the Models}
        \centering
        \begin{tabular}{cccccc}
        \toprule
        & Tiny & Mini & Small & Medium & Base \\
        \midrule
        Capitalization & 0.86 & 0.90 & 0.918 & 0.919 & 0.926 \\
        Punctuation & 0.65 & 0.73 & 0.76 & 0.77 & 0.785 \\
        \bottomrule
        \end{tabular}
        \label{tab:f1Comapre}
    \end{table}

\begin{figure}[htbp]
        \centering
        \includegraphics[scale=0.6]{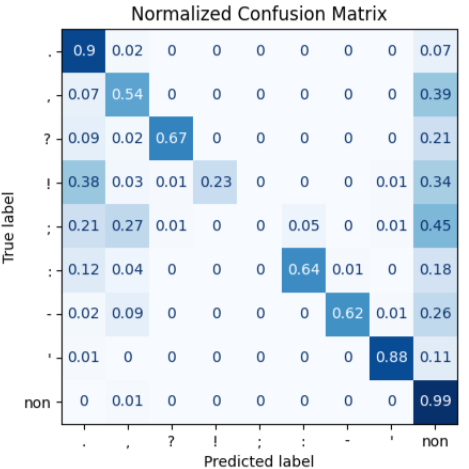}
        \caption{Tiny Punctuation Confusion Matrix}
        \label{fig:cmTinyPunc}
    \end{figure}

\begin{figure}[htbp]
        \centering
        \includegraphics[scale=0.55]{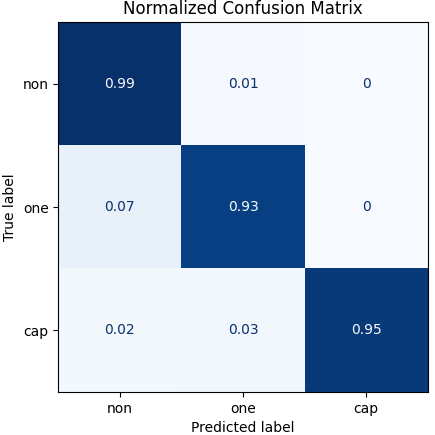}
        \caption{Tiny Capitalization Confusion Matrix}
        \label{fig:cmTinyCap}
    \end{figure}

    We can infer from the confusion matrices of the Tiny models in Figures \textbf{\ref{fig:cmTinyPunc}} and \textbf{\ref{fig:cmTinyCap}} that, while the capitalization model's performance is acceptable, the punctuation model's performance is less reliable. Specifically, the model performs well with periods and apostrophes but struggles with more complex punctuation marks.
    
    The Mini models demonstrated the most significant improvement over the Tiny models, achieving a 0.04 increase in F1 score for capitalization and an 0.08 increase for punctuation, which now accurately predicts most punctuation marks, except for semicolons and exclamation marks.
    


    In Small models considerable improvement over Mini models has been seen, both models are performing effectively in most of the cases, except for semicolons and exclamation marks.

\begin{figure}[htbp]
        \centering
        \includegraphics[scale=0.80]{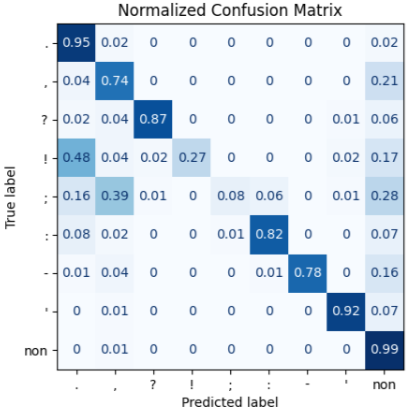}
        \caption{Base Punctuation Confusion Matrix}
        \label{fig:cmBasePunc}
    \end{figure}

\begin{figure}[htbp]
        \centering
        \includegraphics[scale=0.60]{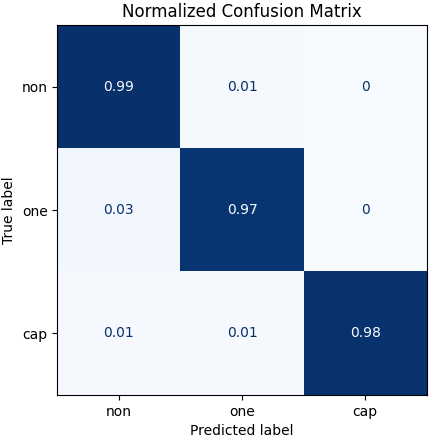}
        \caption{Base Capitalization Confusion Matrix}
        \label{fig:cmBaseCap}
    \end{figure}

    
    Moving from Small to Medium and Base models, minor improvements have been observed, with a 0.926 F1 score for capitalization and 0.785 F1 score for punctuation. The Base model demonstrated the highest performance. The confusion matrices for punctuation and capitalization performance of the base model are presented in Figures \textbf{\ref{fig:cmBasePunc}} and \textbf{\ref{fig:cmBaseCap}}. These values demonstrate an increase for each label compared to the tiny model, indicating a significant improvement. While the tiny model exhibited a lack of accuracy in predicting the semicolon, this score was enhanced with the base model. The performance of the model in relation to the number of punctuation marks is proportional. It is therefore anticipated that the greater the quantity of data utilised, the more effective the performance will be for less common punctuation marks.
    

    
\section{Conclusion and Future Studies}
This study has successfully demonstrated the potential of BERT-based models to effectively correct punctuation and capitalization errors in Turkish text, a language with unique syntactic and grammatical complexities. Through the deployment of five distinct model sizes—Tiny, Mini, Small, Medium, and Base—we have provided a comprehensive analysis that illustrates the relationship between model size, performance, and resource utilization.

The experimental results demonstrate that while larger models, such as the Base model, achieve the highest levels of accuracy, with an F1 score of 0.785 in punctuation and 0.926 in capitalization tasks, smaller models, such as Tiny and Mini, still perform commendably. This balance offers flexibility for applications with limited computational resources or those requiring faster processing times without a substantial sacrifice in performance.

A key finding of this research is the importance of model tuning, particularly the optimization of learning rates and batch sizes, which played a significant role in achieving the best results from each model size. Furthermore, our approach in adapting the BERT architecture for the specific needs of punctuation and capitalization correction has not only improved the readability and quality of Turkish texts but also enhanced the efficiency of text-to-speech systems and automated content editing tools.

As future work, we intend to investigate the integration of these models into real-world applications such as live text editors and automated content generation systems. Additionally, we aim to examine the transferability of our approach to other agglutinative languages, which could expand the impact of our work and provide a foundation for multilingual NLP tools that can adapt to the nuances of various languages.

Additionally, we plan to conduct experiments with significantly larger datasets, examining the impact of corpus size on model performance. This includes a focus on improving the prediction accuracy of less frequently used punctuation marks by strategically increasing their representation in the training data. By adjusting the proportion of paragraphs containing these rarer punctuation marks, we can assess their impact on overall model performance and refine our approach to better handle diverse linguistic scenarios. This effort will enable us to further enhance the robustness and accuracy of our models across a wider range of text types and applications.

In conclusion, the outcomes of this study contribute valuable insights into the development of NLP tools for Turkish and potentially other similar languages. They reaffirm the efficacy of BERT-based models in dealing with complex linguistic tasks, such as punctuation and capitalization correction. The scalability of model sizes provides developers and researchers with a versatile toolkit that can be employed in a range of scenarios, from mobile devices with limited capabilities to powerful cloud-based applications.

\section*{Acknowledgment}
This study was supported by the Scientific and Technological Research Council of Turkey (TUBITAK) Grant No: 124E055.

This work has been supported by Yildiz Technical University Scientific Research Projects Coordination Unit under project number FBA-2024-6070

Research supported with Cloud TPUs from Google's TPU Research Cloud (TRC).


\begin{thebibliography}{00}
\bibitem{dataset_article}
B. Baykara and T. Güngör, 
"Abstractive text summarization and new large-scale datasets for agglutinative languages Turkish and Hungarian," 
\textit{Language Resources and Evaluation}, vol. 1, no. 35, 2022, doi:10.1007/s10579-021-09568-y.

\bibitem{huggingface}
Hugging Face Inc., 
"Hugging Face: The AI community building the future," 2024. [Online]. Available: https://huggingface.co. Accessed: Apr. 20, 2024.

\bibitem{toprak2023developing}
H. T. Kesgin, M. K. Yuce, and M. F. Amasyali, 
"Developing and evaluating tiny to medium-sized Turkish BERT models," 
\textit{arXiv e-prints}, vol. arXiv--2307, 2023.

\bibitem{CRF}
J. D. Lafferty, A. McCallum, and F. C. N. Pereira, 
"Conditional random fields: Probabilistic models for segmenting and labeling sequence data," 
in \textit{Proc. 18th Int. Conf. Machine Learning}, San Francisco, CA, USA: Morgan Kaufmann Publishers Inc., 2001, pp. 282--289.

\bibitem{Viet}
H. Tran, C. V. Dinh, Q. Pham, and B. T. Nguyen, 
"An efficient transformer-based model for Vietnamese punctuation prediction," 
in \textit{Proc. Int. Conf. Industrial, Engineering and Other Applications of Applied Intelligent Systems}, Springer, 2021, pp. 47--58.


\bibitem{bilstm}
"BiLSTM - Papers With Code." [Online]. Available: https://paperswithcode.com/method/bilstm.


\bibitem{mahmoudArtile}
U. Kurt and A. Çayır, 
"Transformer based punctuation restoration for Turkish," 
in \textit{Proc. 8th Int. Conf. Computer Science and Engineering (UBMK)}, IEEE, 2023, pp. 169--174.


\bibitem{bert}
J. Devlin, M.-W. Chang, K. Lee, and K. Toutanova, 
"BERT: Pre-training of deep bidirectional transformers for language understanding," 
\textit{arXiv preprint arXiv:1810.04805}, 2018.

\bibitem{electra}
K. Clark, M.-T. Luong, Q. V. Le, and C. D. Manning, 
"Electra: Pre-training text encoders as discriminators rather than generators," 
\textit{arXiv preprint arXiv:2003.10555}, 2020.

\bibitem{convbert}
Z.-H. Jiang, W. Yu, D. Zhou, Y. Chen, J. Feng, and S. Yan, 
"ConvBERT: Improving BERT with span-based dynamic convolution," 
\textit{Advances in Neural Information Processing Systems}, vol. 33, pp. 12837--12848, 2020.


\bibitem{soudArticle}
T. Alam, A. Khan, and F. Alam, 
"Punctuation restoration using transformer models for high- and low-resource languages," 
in \textit{Proc. 6th Workshop on Noisy User-Generated Text (W-NUT 2020)}, 2020, pp. 132--142.



\bibitem{deneme}
Y. P. Kilic, D. Duygu, and P. Karagoz, 
"Named entity recognition on morphologically rich language: Exploring the performance of BERT with varying training levels," 
in \textit{2020 IEEE Int. Conf. Big Data (Big Data)}, IEEE, 2020, pp. 4613--4619.

\bibitem{ling2023small}
T. Ling, C. Liao, Z. Yu, L. Chen, S. Huang, and Y. Liu, 
"A small and fast BERT for Chinese medical punctuation restoration," 
\textit{arXiv e-prints}, vol. arXiv--2308, 2023.

\bibitem{courtland2020efficient}
M. Courtland, A. Faulkner, and G. McElvain, 
"Efficient automatic punctuation restoration using bidirectional transformers with robust inference," 
in \textit{Proc. 17th Int. Conf. Spoken Language Translation}, 2020, pp. 272--279.


\bibitem{adamW}
I. Loshchilov and F. Hutter, 
"Decoupled weight decay regularization," 
\textit{arXiv preprint arXiv:1711.05101}, 2019. [Online]. Available: https://arxiv.org/abs/1711.05101.








\end{thebibliography}
\end{document}